\documentclass[sigconf]{acmart}
\usepackage{amsfonts}
\usepackage{algorithmic}
\usepackage{textcomp}
\usepackage{url}
\usepackage[inkscapelatex=false]{svg}
\usepackage{multirow}
\usepackage{xspace}
\usepackage{fancyhdr}

\usepackage{multirow,makecell}

\def\BibTeX{{\rm B\kern-.05em{\sc i\kern-.025em b}\kern-.08em
    T\kern-.1667em\lower.7ex\hbox{E}\kern-.125emX}}

\copyrightyear{2026}
\acmYear{2026}
\setcopyright{cc}
\setcctype{by}
\acmConference[AST '26]{7th ACM/IEEE International Conference on Automation of Software Test (AST 2026)}{April 13--14, 2026}{Rio de Janeiro, Brazil}
\acmBooktitle{7th ACM/IEEE International Conference on Automation of Software Test (AST 2026) (AST '26), April 13--14, 2026, Rio de Janeiro, Brazil}
\acmPrice{}
\acmDOI{10.1145/3793654.3793741}
\acmISBN{979-8-4007-2476-3/2026/04}

\begin{document}

\fancypagestyle{firststyle}
{
   \fancyhf{}
   \fancyhead[C]{This is the authors' copy of the paper to appear in the 7th ACM/IEEE International Conference on Automation of Software Test (AST '26).}
   \renewcommand{\headrulewidth}{0pt} 
}

\title[ACT: Automated CPS Testing for Open-Source Robotic Platforms]{ACT: Automated CPS Testing for Open-Source Robotic Platforms}

\newcommand{\pololu}{Pololu 3pi+ 2040 Robot\xspace}

\author{Aditya A. Krishnan}
\affiliation{%
  \institution{Arizona State University}
  \city{Tempe}
  \state{AZ}
  \country{USA}}
\email{aakrishn@asu.edu}
\orcid{0009-0008-9149-4144}

\author{Donghoon Kim}
\affiliation{%
  \institution{Arkansas State University}
  \city{Jonesboro}
  \state{AR}
  \country{USA}}
\email{dhkim@astate.edu}
\orcid{0000-0003-4642-4231}

\author{Hokeun Kim}
\affiliation{%
  \institution{Arizona State University}
  \city{Tempe}
  \state{AZ}
  \country{USA}}
\email{hokeun@asu.edu}
\orcid{0000-0003-1450-5248}






\begin{abstract}

Open-source software for cyber-physical systems (CPS) often lacks robust testing involving robotic platforms, resulting in critical errors that remain undetected.
This is especially challenging when multiple modules of CPS software are developed by various open-source contributors.
To address this gap, we propose \underline{\textbf{A}}utomated \underline{\textbf{C}}PS \underline{\textbf{T}}esting (ACT) that performs automated, continuous testing of open-source software with its robotic platforms, integrated with the open-source infrastructure such as GitHub.
We implement an ACT prototype and conduct a case study on an open-source CPS with an educational robotic platform to demonstrate its capabilities.

\end{abstract}


\copyrightyear{2026}
\acmYear{2026}
\setcopyright{cc}
\setcctype{by}
\acmConference[AST '26]{7th ACM/IEEE International Conference on Automation of Software Test (AST 2026)}{April 13--14, 2026}{Rio de Janeiro, Brazil}
\acmBooktitle{7th ACM/IEEE International Conference on Automation of Software Test (AST 2026) (AST '26), April 13--14, 2026, Rio de Janeiro, Brazil}
\acmPrice{}
\acmDOI{10.1145/3793654.3793741}
\acmISBN{979-8-4007-2476-3/2026/04}

%
%


%

\maketitle
\thispagestyle{firststyle}

\section{Introduction}

Open-source software (OSS) for cyber-physical systems (CPS) and robotics, such as Autoware~\cite{autoware2020} or Robotic Operating System 2 (ROS2)~\cite{ros2_2022} have been widely used for various CPS, including autonomous vehicles, industrial robots, and embedded control systems.
Frequent interactions with the physical world and humans make the safety and correctness of CPS critical; thus, rigorous testing for open-source CPS is sorely needed , contributed by community developers who can unintentionally introduce bugs or errors.

However, the current CPS OSS projects face severe challenges.
For example, the majority of testing infrastructure of CPS OSS relies on simulation~\cite{birchler2025roadmap} or abstraction~\cite{mandrioli2024testing} of the underlying hardware; thus, it does not guarantee correctness of the software when deployed on CPS hardware~\cite{abbaspour2015survey}.
In addition, general challenges in the testing of OSS, such as limited automation~\cite{lin2020test} or insufficient coverage~\cite{kochhar2014empirical}, become more prominent for building CPS testbeds~\cite{NIST_CPS_IoT_Testbed}. 
Another important challenge in open-source CPS testing is the high cost and extensive latency of continuous integration and continuous deployment (CI/CD)~\cite{zampetti2023continuous,singh2019comparison} of CPS OSS.
All in all, these challenges render critical issues and problems in the open-source CPS software undetected and shipped to real-world deployment.

To address this problem, this paper proposes a continuous, cost-effective, \underline{\textbf{A}}utomated \underline{\textbf{C}}PS \underline{\textbf{T}}esting (ACT) platform.
The proposed ACT platform can provide low-cost, low-latency testing of robotic components widely used by open-source CPS software, including testing of status lights, inertial measurement units (IMU), general physical sensors, displays, and actuators such as motors. 

This paper's contributions are threefold: (i) proposal of the ACT approach for open-source CPS software with OSS infrastructure (e.g., GitHub) using a local self-hosted runner~\cite{github2025self} integrated with robotic platforms, (ii) in-depth case study of ACT implementation with real-world CPS OSS called \emph{Lingua Franca (LF)}~\cite{linguafranca2025self} with test cases from educational CPS lab exercises~\cite{embeddedlabs2025_doc} based on LF using a commercial-off-the-shelf robotic hardware, \pololu~\cite{pololu2025self}, and (iii) analysis and diagnosis results demonstrating capabilities and viability of ACT with regards to important topics on testing CPS OSS such as flakiness, testing latency, and stability.


\section{Related Work}
\label{section:related_work}

Testing infrastructure for OSS, such as CI/CD for CPS, has been a bottleneck due to multiple reasons, as identified by
a survey on 20 OSS~\cite{zampetti2022problems}.
One notable bottleneck is resource availability and specific build environments, due to multiple interacting systems that require different software versions in complex CPS.
Another study on industrial CPS projects~\cite{zampetti2023continuous} has shown that unstable testing environments and limited resources are significant challenges in most organizations.
A step towards creating a framework that can be used for testing in CPS would be to have a deterministic and stable build environment paired with improved automated testing that does not hinder already established CI/CD pipelines. 

There has been a rich literature on improving CPS OSS.
A testbed by Shrivastava \textit{et al.}~\cite{shrivastava2017testbed} focuses on testing the timing behavior via various methods using specific testbed setups. Another study~\cite{mokhtarian2024survey} on various robotic testbeds shows the plethora of robots and techniques that are used to test algorithms and robot behaviors.
Online platforms for testing robotic algorithms with physical robots are designed to verify the algorithms that were developed either for educational or prototyping purposes, like the Robot Programming Network (RPN)~\cite{cervera2015robot} and RoboRacer (formerly known as F1TENTH~\cite{koirala2024f1tenth}).
Similarly, this paper also uses educational labs for CPS and embedded systems software as a case study for the proposed approach.

Sadri-Moshkenani~\textit{et al.}~\cite{sadri2022survey} present a comprehensive survey of test-case generation, selection, and prioritization for CPS, characterizing the properties of state-of-the-art techniques.
They also delineate unresolved challenges and outline open research directions across CPS testing. Noticeable among these is the difficulty of designing reliable test oracles for nondeterministic~\cite{sztipanovits2011toward}, continuous behaviors~\cite{briand2016testing}, which often require tolerance bands or acceptable sets rather than single expected values, and the inadequacy of traditional code coverage, which calls for signal and feature-oriented adequacy criteria tailored to continuous, real-time dynamics.

\section{Proposed Automated CPS Testing Approach}
\label{section:proposed_testing_flow}

\begin{figure} 
    \centering
    \includegraphics[width=0.78\linewidth]{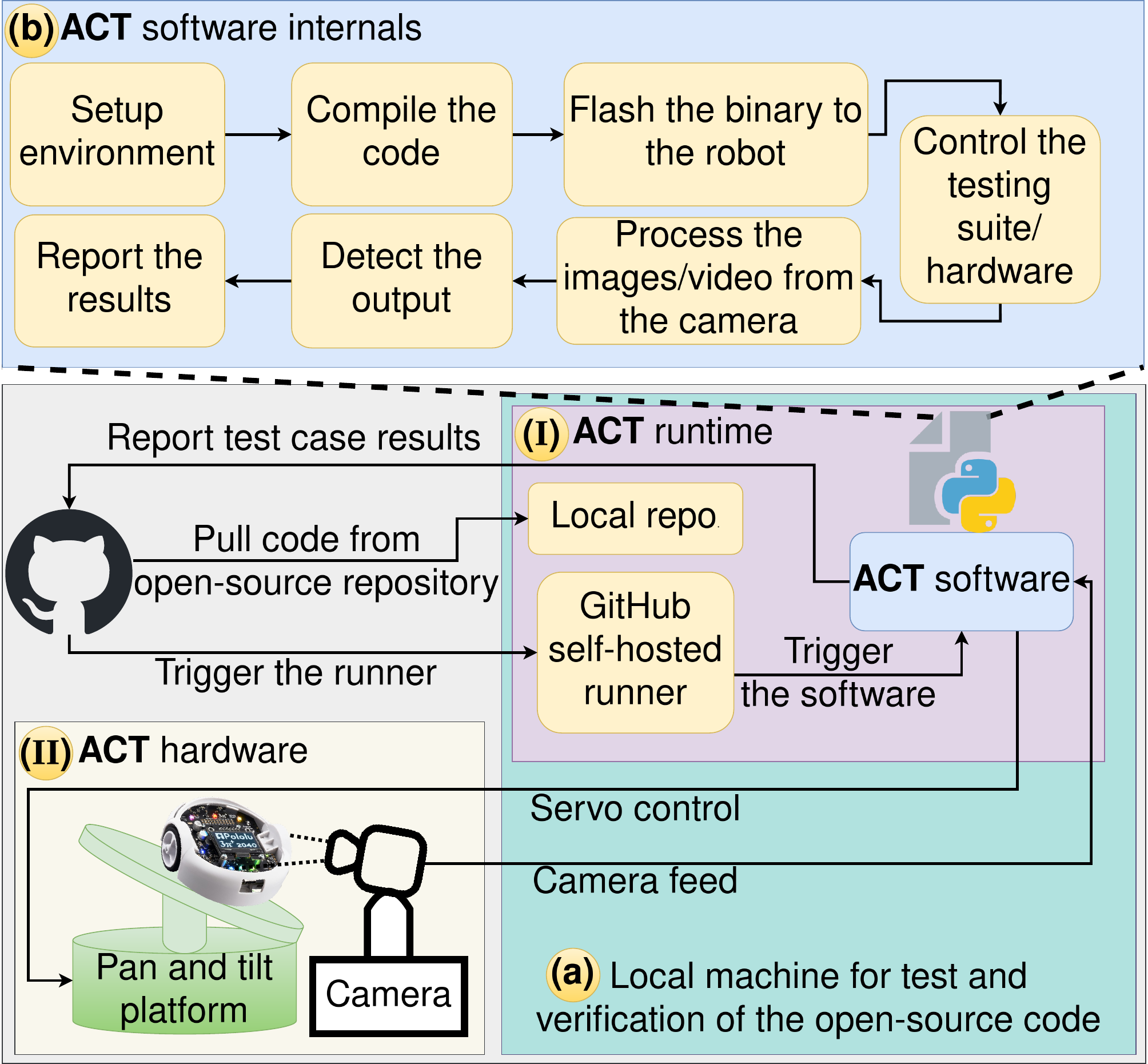}
    \vspace{-10pt}
    \caption{
    Proposed ACT platform for automated testing CPS OSS with robotic hardware;
    (I) ACT runtime with software and a self-hosted runner,
    (II) ACT hardware for testing robotic platform,
    (a) local machine for the self-hosted runner and ACT software, and
    (b) overall flow inside ACT software.
    }
    \vspace{-12pt}
    \label{fig:Testbed}
\end{figure}


\figurename~\ref{fig:Testbed} illustrates the proposed approach, \textbf{ACT}, as a robust automated testing framework for CPS with physical robotic hardware.
Our framework consists of two main components; (i) ACT runtime (\figurename~\ref{fig:Testbed}-I) which includes of the GitHub self-hosted runner~\cite{github2025self} and the software that runs the testing and (ii) ACT hardware (\figurename~\ref{fig:Testbed}-II) that controls the hardware testbed responsible controlling the hardware-based testing mechanisms such as the pan and tilt platform for testing inertial measurement unit (IMU) sensors.

ACT runtime (\figurename~\ref{fig:Testbed}-I) runs on a local machine (\figurename~\ref{fig:Testbed}a) for testing and verification of OSS.
The local machine can be an embedded board (e.g., Raspberry Pi) or a server connected to networks and ACT hardware (\figurename~\ref{fig:Testbed}-II).
Firstly, our GitHub action~\cite{decan2022use} automatically triggers the self-hosted runner when testing is needed, e.g., before merging certain pull requests (PRs).
ACT provides triggering conditions (e.g., merge requests and specific labels) for specific test cases for each peripheral, and these conditions can be configured by the users.
We also provide fine control over which test case to trigger to avoid constant triggering of the test cases.
The workflow will take care of the cloning and compiling of the programs, and the Python software will monitor the test cases and report the values along with the graphs generated. 

We make our ACT runtime open-sourced for follow-up research.\footnote{Available online: https://github.com/lf-lang/act-lf-testbed}

ACT software's internal flow (\figurename~\ref{fig:Testbed}b) starts with the environment setup, cross-compilation, and flashing (i.e., loading) to an executable for the robotic platform.
Then, ACT software performs the testing using the controllable pan and tilt platform, as well as camera modules to acquire test results and measurements.
For each test, ACT software makes verdicts (e.g., pass/fail) and logs detailed reasons, and reports the verdicts/logs to GitHub actions. 

ACT hardware (\figurename~\ref{fig:Testbed}-II)  consists of camera modules, a pan and tilt platform, and actuators to test the bump sensors.
ACT hardware aims to be automated and agnostic of the robotic hardware platform to be tested, although the current configuration assumes only small-scale robotic platforms.
ACT hardware is tailored to perform common testing for robotic platforms, including testing status indicators (e.g., LEDs), IMU sensors (e.g., accelerometers, gyroscopes) using pan and tilt platform, actuators (e.g., motors), physical sensors (e.g., buttons, line/bump sensors), as well as simple displays for the status of the robotic platform in textual format.

ACT is designed to identify both software faults and hardware damage sustained during operation, while providing an evaluation template to address challenges in CPS OSS testing. 
Among issues in OSS testing, flaky tests~\cite{parry2021survey}, which are automated tests that intermittently pass or fail under identical conditions, can be especially problematic. 
ACT aims to adaptively adjust its configurations for the tradeoff among flakiness, latency, and accuracy; pushing for short turnaround at high accuracy tends to raise flakiness, while lowering flakiness often requires longer observation or relaxed accuracy.
Our case study of ACT in Section~\ref{section:case_study} discusses this interplay between flakiness, testing latency, and accuracy with real-world CPS OSS with educational exercises as test cases.





\section{Case Study of Proposed Testing Mechanisms}
\label{section:case_study}


For our case study of ACT, we use an open-source CPS and robotic platform called \emph{Lingua Franca (LF)}~\cite{linguafranca2025self} with an open-source educational materials based on LF, \emph{Embedded Systems Labs}~\cite{embeddedlabs2025_doc}, which are widely used by educational institutions (UC Berkeley, Arizona State University, Southern Illinois University, TU Dresden etc.).
These labs are based on a commercial-off-the-shelf small-scale robot, \pololu~\cite{pololu2025self}, equipped with various sensors and actuators, to provide hands-on experience in embedded systems and CPS design for students.
Starting from introductory labs using status lights (LEDs), each lab builds upon and extends the previous ones to use different sensors, such as an IMU for navigation and orientation detection later on, using motors as actuators.

Our case study focuses on the following five robot components: LED status lights, displays, IMU sensors, bump sensors, and motors. 
These components serve as a representative model of real-world CPS with similar, often more sensitive, safety-critical components, enabling evaluation of both functional correctness and robustness.

%


%

\subsection{LED as Robotic Status Indicators}


Onboard LEDs of robots provide immediate visual feedback on system operation~\cite{liberman2023you}. To verify that the LED is operating correctly, the method applies a color mask and evaluates a fixed region of interest (ROI).
During this part of our case study, the robot is stationary and the camera module is aimed directly at the target LED.

\begin{figure}
    \centering
    \includegraphics[width=0.55\linewidth]{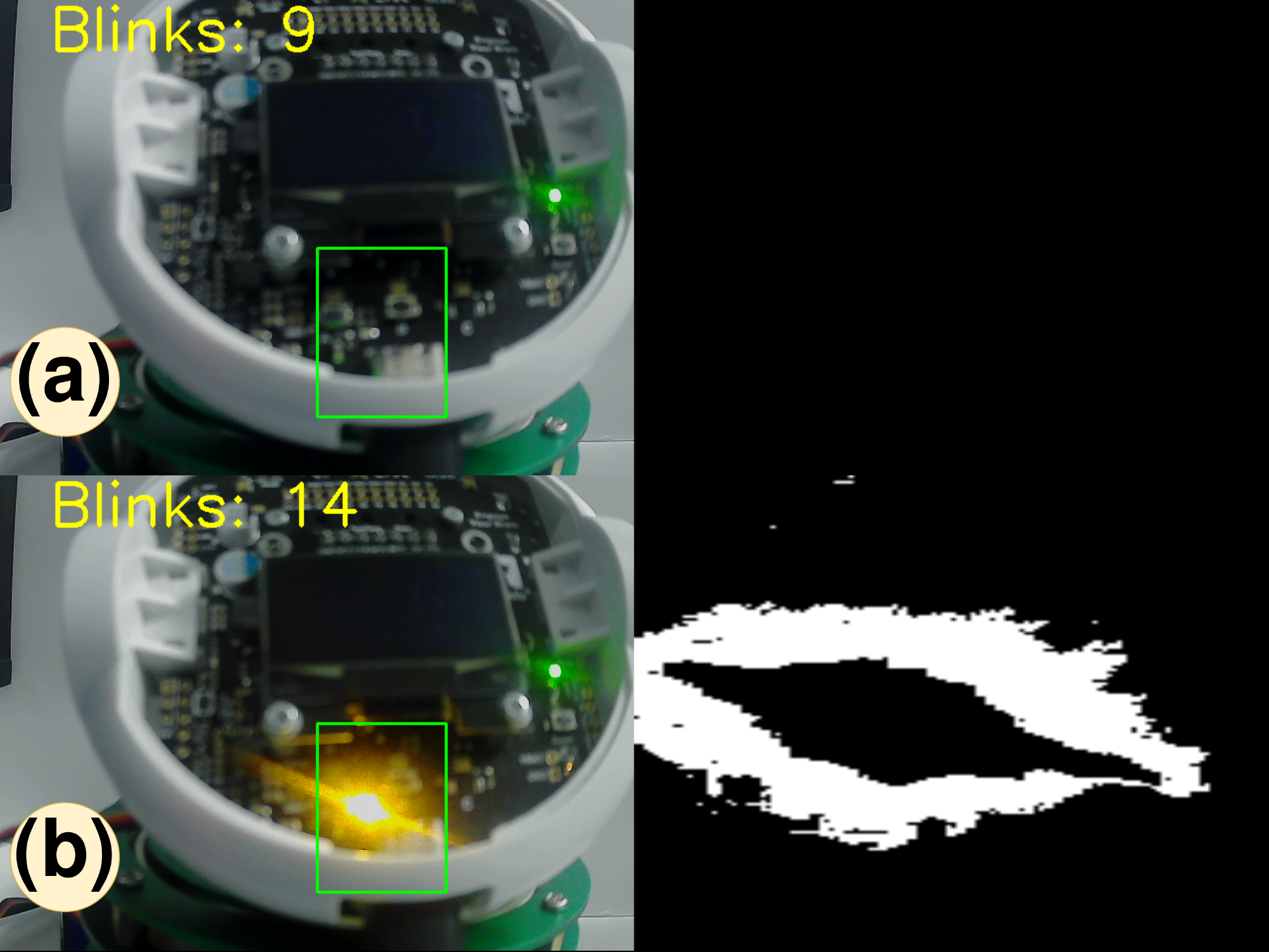}
   \vspace{-10pt}
    \caption{
    Example LED status indicator blink detection using a color filter;
    (a) LED off: no active pixels detected, other LEDs not shown due to filtering.
    (b) LED on: the mask highlights only the LED, filtered by its HSV (hue, saturation and value).
    }
    \label{fig:LED Setup}
\end{figure}

\figurename~\ref{fig:LED Setup} illustrates that the color mask passes only pixels matching the LED’s hue within the ROI, filtering out other light and reducing false positives. When the LED is on (\figurename~\ref{fig:LED Setup}b), it appears as a bright spot.
Counting the frames exceeding the intensity threshold (5 active pixels) within the sampling window and dividing by the window duration yields the LED blink frequency.
%
%
The value converges towards the actual blink frequency as testing time elapses, with the time point ranging between 20 and 30 seconds. When the number of blinks increases, the error margin exponentially decreases. 
Our target error tolerance is 0.5\% of the blinking period, using at least 20 blinks, while maintaining low flakiness with probability at most 1\%.


\begin{figure} 
    \centering
    \includegraphics[width=0.65\linewidth]{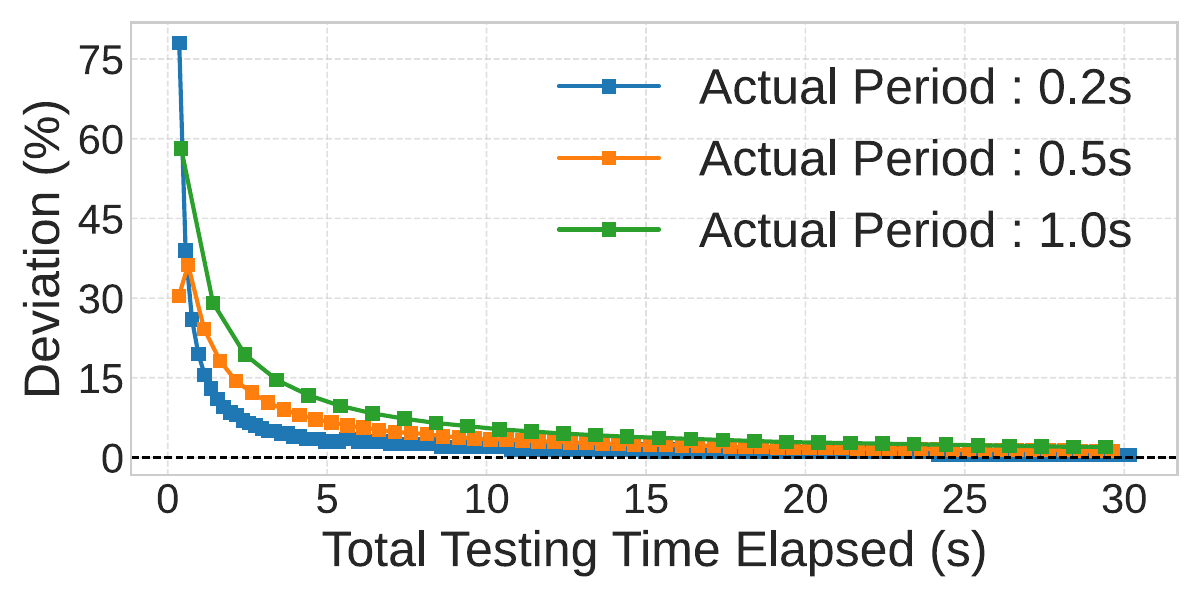}
    \vspace{-10pt}
    \caption{Deviation (in absolute percentage) between actual and measured LED blink periods over elapsed testing time from our case study for different actual blink periods.
    }
    \label{fig:Measured convergence}
    \vspace{-0.5cm}
\end{figure}

\figurename~\ref{fig:Measured convergence} shows how the deviation for each test case converges compared to the actual value. It is evident that a faster blinking sequence can be measured faster and the value converges faster than the other cases where the intervals were relatively higher. The first 5 seconds show the clear distinction in the rate of convergence. The deviation reduces and nears zero in the next 15 to 20 seconds.

\subsection{IMU with Display}

An inertial measurement unit (IMU) is key for robot navigation, e.g., where the robot needs to detect slopes and plateaus~\cite{ahmad2013reviews}. The pitch and roll values are needed for detecting slopes and inclined planes during navigation. To test the IMU, the robot tilt platform is used to keep the robot in a pre-defined incline. The IMU data is displayed for each angle, where we use image processing and optical character recognition (OCR)~\cite{mithe2013optical} to extract the text.
However, this test is not real-time because the display value changes continuously and then does not yield a single definitive result. Instead, a snap of the robot display is taken to process the data.  

\begin{figure}
    \centering
    \includegraphics[width=0.8\linewidth]{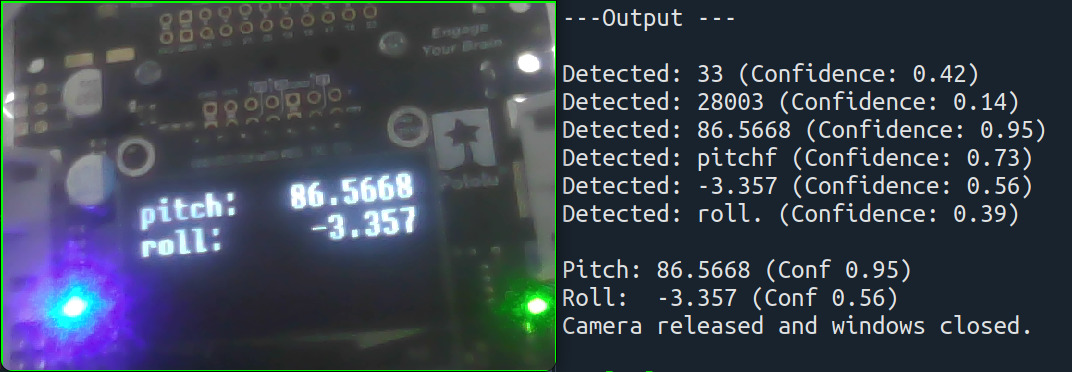}
    \vspace{-10pt}
    \caption{The IMU readings on the robot's display (left) and those measured by ACT's camera (right).}
    \label{fig:IMU Setup}
\end{figure}

\figurename~\ref{fig:IMU Setup} shows the snapshot taken and the detection results with the corresponding confidence of the text in the image.
The measured values denote the values that are processed by the OCR, and the confidence is the score assigned to the actual value.
We use the confidence value from the OCR library, which we leverage to extract only measured values that have high confidence scores. 

\begin{table}
\centering
\caption{Measured IMU sensor data and actual angles under normal/faulty conditions (values are in degrees).}
\footnotesize
\vspace{-10pt}
\begin{tabular}{rr|rr|rr|r}
 & \textbf{Mode} 
 & \begin{tabular}[c]{@{}l@{}}\textbf{Measured}\\\textbf{Pitch}\end{tabular} 
 & \begin{tabular}[c]{@{}l@{}}\textbf{Actual}\\\textbf{Pitch}\end{tabular} 
 & \begin{tabular}[c]{@{}l@{}}\textbf{Measured}\\\textbf{Roll}\end{tabular}
 & \begin{tabular}[c]{@{}l@{}}\textbf{Actual}\\\textbf{Roll}\end{tabular} 
 & \begin{tabular}[c]{@{}l@{}}\textbf{Check}\end{tabular}\\
  \hline
 & Normal & 24.14 & 25.00 & 0.20 &  0.00 & Passed \\
 & Normal & 0.0 & 0.00  & 26.07 & 25.00 & Passed\\
 & Normal & 10.73 & 10.00  & 0.36 & 0.00 & Passed\\
 & Faulty & 217.81 & 20.00  & 2.96 & 0.0 & \textbf{Failed}\\
 & Faulty & 26.56 & 0.0  & 208.10 & 20.00 & \textbf{Failed}\\
 & Faulty & 44.05 & 0.0  & -400.58 & -20.00 & \textbf{Failed}\\
\end{tabular}

\label{tab:imu-data-fault}
\end{table}




\tablename~\ref{tab:imu-data-fault} presents the IMU sensor readings under both normal and fault-injected conditions.
The range for pitch and roll values is from -90 to 90 degrees. The negative sign shows the direction of the orientation about the axes. During a test, we check for either pitch or roll value, which is why one of the two values is always zero. The assigned tolerance is a 5\% deviation to verify if the test can be considered passed or failed. In terms of faults, if the display is faulty, there will be two outcomes. Either there is no output, or there is partial output; both yield a failed test case without the correct value for pitch or roll. In terms of software faults where the calibration is wrong, we can see a significant deviation from the actual value. 

\subsection{Bump Sensors with Display}

\begin{figure}[b]
    \centering
    \includegraphics[width=0.55\linewidth]{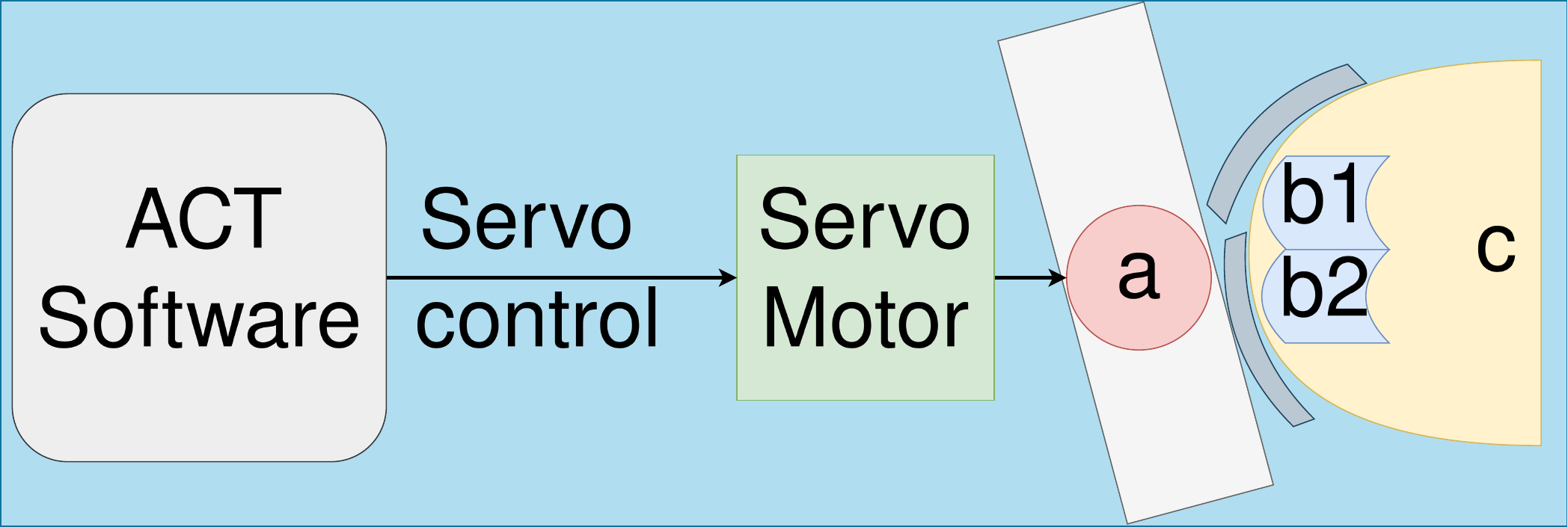}
   \vspace{-10pt}
    \caption{Setup for the bump experiment; a servo motor controlled by ACT software rotates (a) a pivoting block to mechanically press the robot's (b1) right or (b2) left bump sensors on (c) \pololu , generating bump events.}
    \label{fig:Bump Setup}
\end{figure}

\pololu uses optical bump sensors~\cite{mclurkin2014robot}.
Specifically, front bump flaps are monitored by reflective IR sensors, and a sharp change in the reflected signal (after baseline calibration) lets the firmware detect left vs. right bumps. These sensors are critical for obstacle avoidance because the robot does not currently have a camera module installed.
\figurename~\ref{fig:Bump Setup} shows the bump test setup in which ACT software drives a servo motor to physically trigger the robot’s bump sensors. A pivoting block (\figurename~\ref{fig:Bump Setup}a) on the servo alternately taps the right (\figurename~\ref{fig:Bump Setup}b1) or left (\figurename~\ref{fig:Bump Setup}b2) bump on the robot (\figurename~\ref{fig:Bump Setup}c), producing controlled, repeatable bump events. This arrangement enables automated validation of sensor responses and display outputs under normal and fault-injected conditions. 
%

\begin{table*}[t]
\footnotesize
\caption{Experimental results and diagnostics for bump sensors and display under injected faults.}
\label{tab:bump-fault-matrix}
\begin{tabular}{lllllll}
\hline
\multirow{2}{*}{\textbf{\begin{tabular}[c]{@{}l@{}}Fault\\ Classes\end{tabular}}} & \multirow{2}{*}{\textbf{Types of Faults}} & \multicolumn{3}{l}{\textbf{Tested Sensors and Measured (Reported) Results}} & \multirow{2}{*}{\textbf{Display Status}} & \multirow{2}{*}{\textbf{Diagnosis}} \\ \cline{3-5}
 &  & \textbf{\begin{tabular}[c]{@{}l@{}}Left\\ Bump\end{tabular}} & \textbf{\begin{tabular}[c]{@{}l@{}}Right\\ Bump\end{tabular}} & \textbf{\begin{tabular}[c]{@{}l@{}}Both\\ Bumps\end{tabular}} &  &  \\ \hline
\multirow{4}{*}{\textbf{\begin{tabular}[c]{@{}l@{}}Hardware\\ (HW)\end{tabular}}} & Left bump sensor fault & No response to bump & Responds to right bump & \begin{tabular}[c]{@{}l@{}}Responds only to\\ right-bump sensor\end{tabular} & \begin{tabular}[c]{@{}l@{}}Partial or missing\\ data\end{tabular} & \begin{tabular}[c]{@{}l@{}}Left bump sensor\\ hardware fault\end{tabular} \\ \cline{2-7} 
 & Right bump sensor fault & Responds to left bump & No response to bump & \begin{tabular}[c]{@{}l@{}}Responds only to\\ left-bump sensor\end{tabular} & \begin{tabular}[c]{@{}l@{}}Partial or missing\\ data\end{tabular} & \begin{tabular}[c]{@{}l@{}}Right bump sensor\\ hardware fault\end{tabular} \\ \cline{2-7} 
 & Display fault & \begin{tabular}[c]{@{}l@{}}Normal bump\\ detection\end{tabular} & \begin{tabular}[c]{@{}l@{}}Normal bump\\ detection\end{tabular} & \begin{tabular}[c]{@{}l@{}}Normal bump\\ detection\end{tabular} & \begin{tabular}[c]{@{}l@{}}Incorrect or missing\\ output\end{tabular} & \begin{tabular}[c]{@{}l@{}}Display hardware\\ fault\end{tabular} \\ \cline{2-7} 
 & \begin{tabular}[c]{@{}l@{}}Either the display or both of\\ the bump sensors are faulty\end{tabular} & Partial or no output & Partial or no output & Partial or no output & No display output & \begin{tabular}[c]{@{}l@{}}Both sensors are\\ faulty\end{tabular} \\ \hline
\multirow{3}{*}{\textbf{\begin{tabular}[c]{@{}l@{}}Software\\ (SW)\end{tabular}}} & \begin{tabular}[c]{@{}l@{}}Left sensor incorrectly\\ mapped to right\end{tabular} & \begin{tabular}[c]{@{}l@{}}Reports right-bump\\ event\end{tabular} & \begin{tabular}[c]{@{}l@{}}Reports right-bump\\ event\end{tabular} & \begin{tabular}[c]{@{}l@{}}Reports right-bump\\ event\end{tabular} & Normal & \begin{tabular}[c]{@{}l@{}}Left bump sensor\\ mapping error\end{tabular} \\ \cline{2-7} 
 & \begin{tabular}[c]{@{}l@{}}Right sensor incorrectly\\ mapped to left\end{tabular} & \begin{tabular}[c]{@{}l@{}}Reports left-bump\\ event\end{tabular} & \begin{tabular}[c]{@{}l@{}}Reports left-bump\\ event\end{tabular} & \begin{tabular}[c]{@{}l@{}}Reports left-bump\\ event\end{tabular} & Normal & \begin{tabular}[c]{@{}l@{}}Right bump sensor\\ mapping error\end{tabular} \\ \cline{2-7} 
 & \begin{tabular}[c]{@{}l@{}}Both sensors\\ incorrectly-mapped\end{tabular} & \begin{tabular}[c]{@{}l@{}}No bump event\\ reported\end{tabular} & \begin{tabular}[c]{@{}l@{}}No bump event\\ reported\end{tabular} & \begin{tabular}[c]{@{}l@{}}No bump event\\ reported\end{tabular} & No response & \begin{tabular}[c]{@{}l@{}}No response to\\ any bump\end{tabular} \\ \hline
\textbf{Normal} & No fault injected & \begin{tabular}[c]{@{}l@{}}Reports left-bump\\ event\end{tabular} & \begin{tabular}[c]{@{}l@{}}Reports right-bump\\ event\end{tabular} & \begin{tabular}[c]{@{}l@{}}Reports left- and right-\\ bump events\end{tabular} & Normal & \begin{tabular}[c]{@{}l@{}}Sensors operating\\ correctly\end{tabular} \\ \hline
\end{tabular}
\end{table*}

\tablename~\ref{tab:bump-fault-matrix} summarizes the fault injection results for the bump sensors and the display under both hardware and software fault scenarios.
Hardware faults such as left or right sensor failures cause asymmetric or missing bump responses, while display faults lead to incorrect or absent visual feedback. Software mapping errors, in contrast, produce inverted or duplicated bump events even when the sensors physically function correctly. When both hardware and software faults occur, the system shows no response or only a partial response. This is easy to distinguish from normal operation, where both left and right bump events are correctly detected and displayed. These results demonstrate that ACT can systematically identify and differentiate between hardware and software-level faults based on sensor behavior and display output consistency.

%

\subsection{Motors as Actuators}

\begin{figure}
    \centering
    \includegraphics[width=0.65\linewidth]{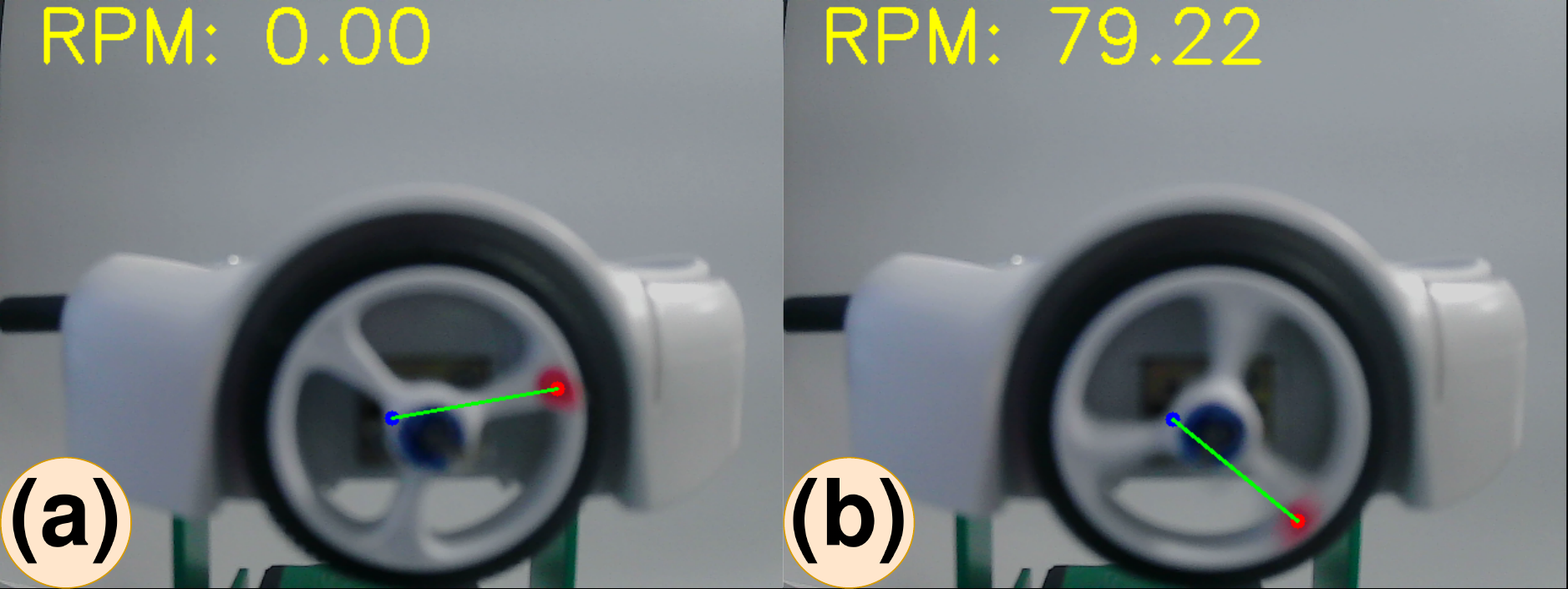}
    \vspace{-10pt}
    \caption{Wheel-motion tracking for RPM estimation using a video feed with
    the (a) wheel stationary (RPM $\approx 0$) and the (b) wheel rotating at RPM $\approx 79$.
    }
    
    \label{fig:Motor Setup}
    \vspace{-10 pt}
\end{figure}

Motors are one of the most important actuators in robots~\cite{kober2009learning}.
\pololu has two DC gearmotors~\cite{pololu2378} that drive the wheels as actuators, which is the critical parts of robots.  
At 6 V without load, the drive motor can reach about 720 RPM, but its peak efficiency is approximately 33\% near 510 RPM~\cite{pololu2378}. The excessive motor speed can damage robot hardware; operation near the maximum speed should be avoided. 
We calibrated the RPM based on the power input manually, where the input values of the duty cycle were 0.05, 0.08, and 0.1 for 33, 60, and 80 RPMs, respectively.

\figurename~\ref{fig:Motor Setup} shows the output when the video feed is processed.
We use the camera to test the robot motors and measure speed.
The robot wheel would be marked with the two dots, one to determine the center of the wheel and the other to determine the edge of one spoke.
A fixed blue point and a red moving marker are tracked in screen captures; the red marker's displacement relative to the blue point yields the motor RPM.
Then we use the filter to mask out the other colors and only use the dots that would be highlighted in the processed image frame.
The relative angle of the spoke with respect to the center gives the number of rotations that were measured in that time period.
Dividing the absolute value of that number by the elapsed time gives the approximate RPM of the robot wheel. 

\begin{figure} 
    \centering
    \includegraphics[width=1\linewidth]{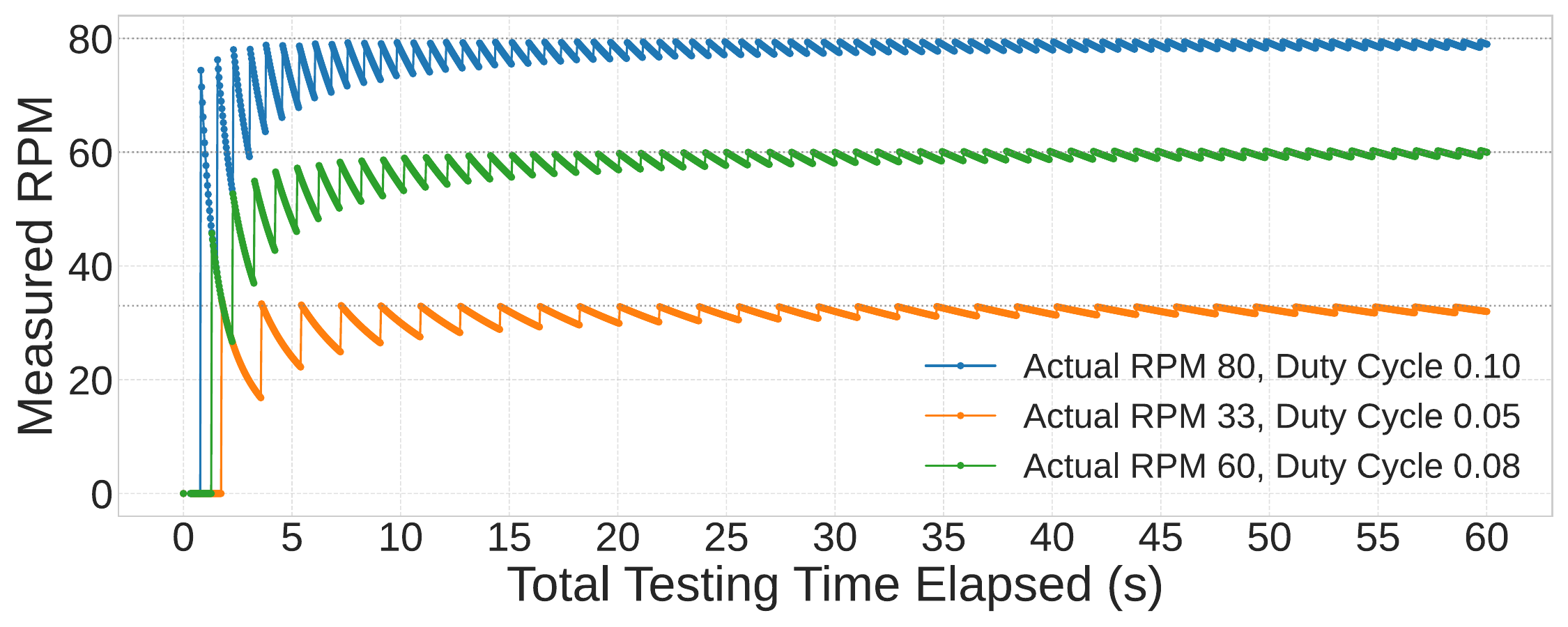}
    \vspace{-25 pt}
    \caption{Motor RPM detection results.
    }
    \label{fig:motor_speed}
     \vspace{-10 pt}
\end{figure}

\figurename~\ref{fig:motor_speed} shows RPM estimates for a suspended wheel, so effects from robot weight and rolling friction are eliminated.
Early readings fluctuate because RPM is computed from the relative angle over very short intervals, where a few rotations can inflate the instantaneous value. 
We deem the RPM estimate stable when, over the last 30\,s, the coefficient of variation is below 3\%, the range is within $\max(\pm 2\,\mathrm{RPM},\, \pm 3\%\ \text{of the mean})$, and the linear trend magnitude is below 0.02 RPM. Empirically, variance decreases as the observation window grows and the estimate converges to the actual RPM values (intended by software); these criteria are typically satisfied after roughly 40--60\,s, so we adopt a 60\,s window to reduce flakiness.


\section{Conclusion}
\label{section:conclusion}

Our concrete case study using real-world CPS OSS~\cite{linguafranca2025self} and its educational labs~\cite{embeddedlabs2025_doc} demonstrates the robust testing capabilities and viability of ACT, our proposed automated CPS testing framework.
ACT provides testing mechanisms for commercial-off-the-shelf robots' sensors/actuators with a specially designed testbed, which can probe the sensor and monitor the output in real time using the physical hardware.
ACT hardware captures the timing in real time and tests the hardware against noise and physical uncertainties.
ACT has the potential to provide better coverage of edge cases and ensure the robustness of CPS OSS.

\section*{Acknowledgment}
This work was supported in part by the National Science Foundation (NSF) under Award No. POSE-2449200 (An Open-Source Ecosystem to Coordinate Integration of Cyber-Physical Systems), OIA-1946391, and OIA-2445877. 

\bibliographystyle{ACM-Reference-Format}
\bibliography{references}
\end{document}